\documentclass{ueacmpstyle}

\usepackage[round]{natbib}

\begin{document}

\title{Survey on Feature Selection}

\author{
Tarek Amr, Beatriz de La Iglesia\\
School of Computing Sciences\\ 
University of East Anglia\\
Norwich Research Park, Norwich, NR4 7TJ, UK
}

\maketitle

\begin{abstract}
Feature selection plays an important role in the data mining process. It is needed to deal with the excessive number of features, which can become a computational burden on the learning algorithms. It is also necessary, even when computational resources are not scarce, since it improves the accuracy of the machine learning tasks, as we will see in the upcoming sections. In this review, we discuss the different feature selection approaches, and the relation between them and the various machine learning algorithms.
\end{abstract}

\section{Introduction}
According to \cite{Dunham-2002}, machine learning tasks can be seen as \emph{predictive} or \emph{descriptive} ones. Classification is an example of predictive models. \cite{Friedman-1997} described it as a model where discrete output values (class labels) are learnt from the different variables (features) of the input data. Clustering, on the other hand, is categorised by \cite{Dunham-2002} as a descriptive task. The features of the input data are used to categorize it without supervised training. In both cases, the choice of the feature-set plays an important role in the performance of the data mining problem. \cite{Liu-2010} listed three advantages for removing irrelevant and redundant features: it makes the data mining task more efficient, improves its accuracy and simplifies the inferred model, making it more comprehensible.

For an accurate classifier, it is needed to reduce both bias and variance of the model~\citep{Friedman-1997}. As described by \cite{Domingos-1999}, the bias is a systematic error that occurs when inferring a more generalized model for the data, hence increasing the training data will not improve it. Variance, on the other hand, results when the model tries to cope with the variations of the noisy data sample. Increasing the sample size in this case can balance the effect of the noise and reduce the variance accordingly. Nevertheless, \cite{Friedman-1997} stresses that during the training process, the more sensitive the model is to the training data, the lower the bias in exchange for a higher variance, and vice versa. This is known as \emph{``bias-variance trade-off"}. Hence, as noted by \cite{Kohavi-1997}, classifiers faced with limited data has to find an optimum point where they can actually estimate the statistical distribution of fewer features (variance reduction) versus less accurate estimation of more features (bias reduction); ergo, \cite{Munson-2009} summarized the feature selection process as the process of finding the best bias-variance trade-off point.

When it comes to unsupervised learning algorithms, such as clustering, \cite{Janecek-2008} highlighted that the problem with high dimensional data (more features) is that it makes the proximity measures between the records more uniform, hence metrics such as distance and density become harder to obtain.

In the next sections we explain the different feature selection approaches.

\section{The selection process}
\label{sec:selection}

In its simplest form, the feature selection process can evaluate individual features and rank them based on their correlation with class labels~\citep{Yu-2004}. However, \cite{Hall-1999} reported that studies had proven a good feature subset to be the one whose features are not correlated to each other, besides them being correlated to class labels. Hence, they are better evaluated as subset rather than individually. \cite{Liu-2010} summarized subset feature selection process into three main steps:

\begin{itemize}
 \item Search: Generating a subset of the available features to be evaluated.
 \item Evaluation: Evaluating the utility of the generated subset  
 \item Stop: Deciding whether to stop or continue the search till a stopping criterion is reached
\end{itemize}

For N-dimension feature-space, there are $2^N$ possible subsets. Thus, the generation step uses different approaches to traverse the available subsets. Additionally, instead of searching within all possible subsets, it can stop after reaching certain number of features or iterations, or when an optimum subset is reached according to the evaluation step~\citep{Dash-2003}. 

\cite{John-1994} listed two search algorithms. The \emph{forward selection} algorithms starts with an empty set and keeps adding features, while the \emph{backward elimination} starts with all the features and keeps on removing ones. \cite{Pintelas-2004} explained that in the two algorithms, once a feature is added it cannot be removed and vice versa. Thus, they are described by \cite{Yang-1998, Hall-1999} as \emph{greedy hill-climbing algorithms}, where they assume monotonicity of the whole process. \cite{Kohavi-1997} added that dealing with smaller subset in the beginning makes the forward selection faster than the backward elimination algorithm and can reach relatively fewer features more quickly, yet the latter usually selects more interactive features. This makes Forward Selection preferable with high dimensional data. Generally, hill-climbing algorithms might get caught in local minima and fail to include useful features, or exclude irrelevant ones.

\cite{Hall-1999} mentioned another algorithm; \emph{best first search}. Unlike the hill climbing algorithms, at each step it generates all possible moves and allow for backtracking once the path it is traversing is not adding any improvement. \cite{Kohavi-1997} highlighted the importance of having a stoppage criterion, to limit the generation of possible moves at each step, in order to prevent the algorithm from traversing the entire search space.

\cite{Yang-1998} argued against the monotonicity assumption, presenting \emph{genetic algorithms} as an alternative to escape the local minimas. \cite{Pintelas-2004} explained it as follows: The features are represented as a binary string where zeros represent the absence of features and ones represent their presence. Genetic operations such as mutation (adding or removing a feature by reversing the value of the bit representing it) and crossover (combining two subsets together) take place on the strings and better feature subsets have more chance to produce newer subsets via more mutations and crossovers.

One idea proposed by \cite{Yu-2004}, is to start with individual feature selection first, to eliminating irrelevant features Then subset selection is performed later to remove redundant features. By decoupling the two processes, they downsize the search space for the subset selection, hence improving its performance. However, this contradicts with what {Kohavi-1997} warned of, where an irrelevant feature on its own can still form, among others, an optimal subset. 

In each iteration, the generated subset has to be evaluated. \cite{Dash-2003} explained that this step compares the new subset with the previously acquired ones, or with predefined optimum threshold to decide (i) whether the new subset should replace the previous best subset, and also (ii) whether a stopping criterion has been reached to prevent the algorithm from doing exhaustive search.

The way evaluation is done is what subdivides feature selection into to two main categories: \emph{filters} and \emph{wrappers}~\citep{Hall-1999, Liu-2010}. The two approaches are discussed in the next section.

\section{Filters and Wrappers}
\label{sec:filters-wrappers}

Filters and wrappers are two evaluation strategies. In filters, individual features or subsets are evaluated independently of the learning algorithms, while wrappers use the learning algorithm to evaluate feature subsets~\citep{Estevez-2009}. 

\cite{Gheyas-2010} listed some filter methods such as: mutual information~\citep{Lewis-1992, Peng-2005, Estevez-2009}, chi-square test~\citep{Jin-2006} and Pearson correlation coefficients~\citep{Biesiada-2007}. 

For individual selection, \cite{Lewis-1992} measures the mutual information (MI) between each feature and the target class label. Then features are ranked accordingly, selecting the top n features. \cite{Hamming-1986} stated the following equation to calculate the MI between two variables, $A = [a_1, a-2, .. a_n]$ and $B = [b_1, b_2, .. b_n]$:

\[ I(A, B) = \Sigma_i \Sigma_j Pr(a_i, b_j) \log{ \frac{Pr(a_i, b_j)}{Pr(a_i) * Pr(b_j)} }  \]    

It is clear from the previous equation that for features with equal conditional probability with a class, the rare ones get higher scores than common ones~\citep{Yang-1997}. On contrary, \cite{Yang-1997} added that Chi-Squared values are normalized, but it is not suitably for rare features, since they hardly follow $X^2$ distribution. Linear correlation coefficient is another option, however \cite{Yu-2004, Gomez-2009} warned that the assumption of linear relation between variables and classes is not usually valid; therefore, MI is still widely used. 


It worth mentioning here that some papers, such as \cite{Yang-1997}, discriminate between Information Gain (IG) and Mutual Information (MI), however \citet[p.~21]{Cover-2006} showed that the MI formula mentioned above is the same one referred to by \cite{Quinlan-1986, Hall-1999} as IG.

Traditionally, filter methods select features individually. One idea is to calculate MI between class labels and subsets instead of individual features. However, \cite{Ding-2005} explained that the more variables in our joint probabilities, the harder it is for our limited sample to cover the multivariate density. Hence, they proposed a ``minimal-redundancy-maximal-relevance" (mRMR) formula, which accounts for both inter-features and feature-to-class MI. Both \cite{Peng-2005} and \cite{Estevez-2009} built on this idea. Similarly, \cite{Torkkola-2003} proposed the use of Renyi's entropy as an alternative to \cite{Shannon-1949}'s entropy to solve the multivariate issue, whereas Markov blanket, presented \cite{Koller-1996}, is one other solution.

The absence of target labels in unsupervised learning encouraged \cite{He-2006} to use Laplacian Score (LS). LS assumes that a relevant feature is the one where neighbouring records across the whole feature space are also close across this feature vector~\citep{He-2006}. They added that LS yields to Fisher Criterion Score (FCS) when target labels are available. \cite{Yan-2007, Fu-2008} highlighted that these methods assume classes to be normally distribution across the data-space. Hence, \cite{Dhir-2009} proposed a hybrid measurement based on FCS and MI 

Although filters are normally faster than wrappers, \cite{John-1994} warned that it doesn't take into its consideration the biases of the learning algorithm during subset selection, after showing the wrappers effectiveness. Wrapper methods use the learning algorithm, during the evaluation step, to determine the utility of a certain features subset based on the algorithm's accuracy while using that specific subset~\citep{John-1994}. \cite{Hall-1999} added that the training data is usually divided into folds and accuracy is determined using cross-validation. Compared to filters, \cite{Gheyas-2010} stated that wrapper's effectiveness comes at the expense of their computational cost. Because of their cost, \cite{Pintelas-2004} noticed that the forward selection algorithms (mentioned earlier) might be more common with wrappers, even if it is less effective than the backward selection. \cite{He-2006} also added that wrappers are common in unsupervised learning scenarios, since filters, other than Laplacian Score, usually rely on class labels to calculate the correlations between features and those labels.

\section{FS and Learning Algorithms}

We have stated earlier that good features are not only the ones highly correlated with the target class, but also the ones not correlated with each other. \cite{Kohavi-1997} highlighted that the accuracy of instance-based algorithms is vulnerable to the former, while Naive-Bayes is more robust when faced with the former yet vulnerable to the latter. Nearest neighbour (NN) algorithm is an examples of instance based learning. \cite[p.~116]{Witten-2005} added that the adoption of k-NN, where ($k>1$), can smooth the effect of noisy data a bit, hence variance.

\cite{Lal-2006} remarked that some learning algorithms, such as decision trees (DT), select relevant features implicitly. \cite{Guyon-2003} added that in those embedded methods of feature selection, the selection process takes place during the training phase, rather than in preprocessing step. Nevertheless, decision tree still need earlier feature selection, as noted by \cite{Kohavi-1997}. 

Additionally, \cite{Kohavi-1997} noticed in their experiments that different search algorithms work better with different learning algorithms as well as datasets. Similarly, experiments by \cite{Hua-2009} proved different feature-selection methods to give various accuracy across different sample sizes and data nature.  

\section{Conclusion}
\label{sec:conclusion}

We have seen that wrappers are generally more accurate then filters, yet the latter is more computational efficient. Similar trade-offs exist between selecting the features individually or as a subset, as well as between the different search algorithms. However, experiments showed that the nature of the dataset, the robustness of the classifier and the nature of the learning problem dictates our choices between those trade-offs. Additionally, there are efforts being put to make filter methods suitable to subset selection and unsupervised learning scenarios.
 


\bibliographystyle{plainnat}
\bibliography{fs.bib}{}

\begin{thebibliography}{36}
\providecommand{\natexlab}[1]{#1}
\providecommand{\url}[1]{\texttt{#1}}
\expandafter\ifx\csname urlstyle\endcsname\relax
  \providecommand{\doi}[1]{doi: #1}\else
  \providecommand{\doi}{doi: \begingroup \urlstyle{rm}\Url}\fi

\bibitem[Biesiada and Duch(2007)]{Biesiada-2007}
J.~Biesiada and W.~Duch.
\newblock Feature selection for high-dimensional data, a pearson redundancy
  based filter.
\newblock \emph{Computer Recognition Systems 2}, pages 242--249, 2007.

\bibitem[Cover and Thomas(2006)]{Cover-2006}
T.M. Cover and J.A. Thomas.
\newblock \emph{Elements of information theory}.
\newblock Wiley-interscience, 2006.

\bibitem[Dash and Liu(2003)]{Dash-2003}
M.~Dash and H.~Liu.
\newblock Consistency-based search in feature selection.
\newblock \emph{Artificial intelligence}, 151\penalty0 (1):\penalty0 155--176,
  2003.

\bibitem[Dhir and Lee(2009)]{Dhir-2009}
C.~Dhir and S.~Lee.
\newblock Hybrid feature selection: combining fisher criterion and mutual
  information for efficient feature selection.
\newblock \emph{Advances in Neuro-Information Processing}, pages 613--620,
  2009.

\bibitem[Ding and Peng(2005)]{Ding-2005}
C.~Ding and H.~Peng.
\newblock Minimum redundancy feature selection from microarray gene expression
  data.
\newblock \emph{Journal of bioinformatics and computational biology},
  3\penalty0 (02):\penalty0 185--205, 2005.

\bibitem[Domingos(1999)]{Domingos-1999}
P.~Domingos.
\newblock The role of occam's razor in knowledge discovery.
\newblock \emph{Data Mining and Knowledge Discovery}, 3\penalty0 (4):\penalty0
  409--425, 1999.

\bibitem[Dunham(2002)]{Dunham-2002}
Margaret~H. Dunham.
\newblock \emph{Data Mining: Introductory and Advanced Topics}.
\newblock Prentice-Hall, 2002.
\newblock ISBN 0-13-088892-3.

\bibitem[Est{\'e}vez et~al.(2009)Est{\'e}vez, Tesmer, Perez, and
  Zurada]{Estevez-2009}
P.A. Est{\'e}vez, M.~Tesmer, C.A. Perez, and J.M. Zurada.
\newblock Normalized mutual information feature selection.
\newblock \emph{Neural Networks, IEEE Transactions on}, 20\penalty0
  (2):\penalty0 189--201, 2009.

\bibitem[Friedman(1997)]{Friedman-1997}
J.H. Friedman.
\newblock On bias, variance, 0/1—loss, and the curse-of-dimensionality.
\newblock \emph{Data mining and knowledge discovery}, 1\penalty0 (1):\penalty0
  55--77, 1997.

\bibitem[Fu et~al.(2008)Fu, Yan, and Huang]{Fu-2008}
Y.~Fu, S.~Yan, and T.S. Huang.
\newblock Classification and feature extraction by simplexization.
\newblock \emph{Information Forensics and Security, IEEE Transactions on},
  3\penalty0 (1):\penalty0 91--100, 2008.

\bibitem[Gheyas and Smith(2010)]{Gheyas-2010}
I.A. Gheyas and L.S. Smith.
\newblock Feature subset selection in large dimensionality domains.
\newblock \emph{Pattern Recognition}, 43\penalty0 (1):\penalty0 5--13, 2010.

\bibitem[G{\'o}mez-Verdejo et~al.(2009)G{\'o}mez-Verdejo, Verleysen, and
  Fleury]{Gomez-2009}
V.~G{\'o}mez-Verdejo, M.~Verleysen, and J.~Fleury.
\newblock Information-theoretic feature selection for functional data
  classification.
\newblock \emph{Neurocomputing}, 72\penalty0 (16):\penalty0 3580--3589, 2009.

\bibitem[Guyon and Elisseeff(2003)]{Guyon-2003}
I.~Guyon and A.~Elisseeff.
\newblock An introduction to variable and feature selection.
\newblock \emph{The Journal of Machine Learning Research}, 3:\penalty0
  1157--1182, 2003.

\bibitem[Hall(1999)]{Hall-1999}
M.A. Hall.
\newblock \emph{Correlation-based feature selection for machine learning}.
\newblock PhD thesis, The University of Waikato, 1999.

\bibitem[Hamming(1986)]{Hamming-1986}
R.W. Hamming.
\newblock \emph{Coding and information theory}.
\newblock Prentice-Hall, Inc., 1986.

\bibitem[He et~al.(2006)He, Cai, and Niyogi]{He-2006}
X.~He, D.~Cai, and P.~Niyogi.
\newblock Laplacian score for feature selection.
\newblock \emph{Advances in Neural Information Processing Systems},
  18:\penalty0 507, 2006.

\bibitem[Hua et~al.(2009)Hua, Tembe, and Dougherty]{Hua-2009}
J.~Hua, W.D. Tembe, and E.R. Dougherty.
\newblock Performance of feature-selection methods in the classification of
  high-dimension data.
\newblock \emph{Pattern Recognition}, 42\penalty0 (3):\penalty0 409--424, 2009.

\bibitem[Janecek et~al.(2008)Janecek, Gansterer, Demel, and
  Ecker]{Janecek-2008}
A.G.K. Janecek, W.N. Gansterer, M.~Demel, and G.F. Ecker.
\newblock On the relationship between feature selection and classification
  accuracy.
\newblock In \emph{JMLR: Workshop and Conference Proceedings}, volume~4, pages
  90--105. Citeseer, 2008.

\bibitem[Jin et~al.(2006)Jin, Xu, Bie, and Guo]{Jin-2006}
X.~Jin, A.~Xu, R.~Bie, and P.~Guo.
\newblock Machine learning techniques and chi-square feature selection for
  cancer classification using sage gene expression profiles.
\newblock \emph{Data Mining for Biomedical Applications}, pages 106--115, 2006.

\bibitem[John et~al.(1994)John, Kohavi, Pfleger, et~al.]{John-1994}
G.H. John, R.~Kohavi, K.~Pfleger, et~al.
\newblock Irrelevant features and the subset selection problem.
\newblock In \emph{Proceedings of the eleventh international conference on
  machine learning}, volume 129, pages 121--129. San Francisco, 1994.

\bibitem[Kohavi and John(1997)]{Kohavi-1997}
R.~Kohavi and G.H. John.
\newblock Wrappers for feature subset selection.
\newblock \emph{Artificial intelligence}, 97\penalty0 (1):\penalty0 273--324,
  1997.

\bibitem[Koller and Sahami(1996)]{Koller-1996}
D.~Koller and M.~Sahami.
\newblock Toward optimal feature selection.
\newblock 1996.

\bibitem[Lal et~al.(2006)Lal, Chapelle, Weston, and Elisseeff]{Lal-2006}
T.~Lal, O.~Chapelle, J.~Weston, and A.~Elisseeff.
\newblock Embedded methods.
\newblock \emph{Feature Extraction}, pages 137--165, 2006.

\bibitem[Lewis(1992)]{Lewis-1992}
D.D. Lewis.
\newblock Feature selection and feature extraction for text categorization.
\newblock In \emph{Proceedings of the workshop on Speech and Natural Language},
  pages 212--217. Association for Computational Linguistics, 1992.

\bibitem[Liu et~al.(2010)Liu, Motoda, Setiono, and Zhao]{Liu-2010}
H.~Liu, H.~Motoda, R.~Setiono, and Z.~Zhao.
\newblock Feature selection: An ever evolving frontier in data mining.
\newblock In \emph{Proc. The Fourth Workshop on Feature Selection in Data
  Mining}, volume~4, pages 4--13, 2010.

\bibitem[Munson and Caruana(2009)]{Munson-2009}
M.~Munson and R.~Caruana.
\newblock On feature selection, bias-variance, and bagging.
\newblock \emph{Machine Learning and Knowledge Discovery in Databases}, pages
  144--159, 2009.

\bibitem[Peng et~al.(2005)Peng, Long, and Ding]{Peng-2005}
H.~Peng, F.~Long, and C.~Ding.
\newblock Feature selection based on mutual information criteria of
  max-dependency, max-relevance, and min-redundancy.
\newblock \emph{Pattern Analysis and Machine Intelligence, IEEE Transactions
  on}, 27\penalty0 (8):\penalty0 1226--1238, 2005.

\bibitem[Pintelas(2004)]{Pintelas-2004}
S.B.K.P.E. Pintelas.
\newblock On the selection of classifier-specific feature selection algorithms.
\newblock In \emph{Proceedings of International Conference on Intelligent
  Knowledge Systems (IKS-2004)}, 2004.

\bibitem[Quinlan(1986)]{Quinlan-1986}
J.R. Quinlan.
\newblock Induction of decision trees.
\newblock \emph{Machine learning}, 1\penalty0 (1):\penalty0 81--106, 1986.

\bibitem[Shannon et~al.(1949)Shannon, Weaver, Blahut, and Hajek]{Shannon-1949}
C.E. Shannon, W.~Weaver, R.E. Blahut, and B.~Hajek.
\newblock \emph{The mathematical theory of communication}, volume 117.
\newblock University of Illinois press Urbana, 1949.

\bibitem[Torkkola(2003)]{Torkkola-2003}
K.~Torkkola.
\newblock Feature extraction by non parametric mutual information maximization.
\newblock \emph{The Journal of Machine Learning Research}, 3:\penalty0
  1415--1438, 2003.

\bibitem[Witten and Frank(2005)]{Witten-2005}
I.H. Witten and E.~Frank.
\newblock \emph{Data Mining: Practical machine learning tools and techniques}.
\newblock Morgan Kaufmann, 2005.

\bibitem[Yan et~al.(2007)Yan, Xu, Zhang, Zhang, Yang, and Lin]{Yan-2007}
S.~Yan, D.~Xu, B.~Zhang, H.J. Zhang, Q.~Yang, and S.~Lin.
\newblock Graph embedding and extensions: A general framework for
  dimensionality reduction.
\newblock \emph{Pattern Analysis and Machine Intelligence, IEEE Transactions
  on}, 29\penalty0 (1):\penalty0 40--51, 2007.

\bibitem[Yang and Honavar(1998)]{Yang-1998}
J.~Yang and V.~Honavar.
\newblock Feature subset selection using a genetic algorithm.
\newblock \emph{Intelligent Systems and Their Applications, IEEE}, 13\penalty0
  (2):\penalty0 44--49, 1998.

\bibitem[Yang and Pedersen(1997)]{Yang-1997}
Y.~Yang and J.O. Pedersen.
\newblock A comparative study on feature selection in text categorization.
\newblock In \emph{MACHINE LEARNING-INTERNATIONAL WORKSHOP THEN CONFERENCE-},
  pages 412--420. MORGAN KAUFMANN PUBLISHERS, INC., 1997.

\bibitem[Yu and Liu(2004)]{Yu-2004}
L.~Yu and H.~Liu.
\newblock Efficient feature selection via analysis of relevance and redundancy.
\newblock \emph{The Journal of Machine Learning Research}, 5:\penalty0
  1205--1224, 2004.

\end{thebibliography}

\end{document}